\documentclass[10pt,twocolumn,letterpaper]{article}

\usepackage[pagenumbers]{iccv} % To force page numbers, e.g. for an arXiv version
\usepackage{soul}
\usepackage[normalem]{ulem} % For \sout{} (better than \st{})
\usepackage{algorithm}
\usepackage{algpseudocode}

\definecolor{iccvblue}{rgb}{0.21,0.49,0.74}
\definecolor{wacvblue}{rgb}{0.21,0.49,0.74}
\usepackage[pagebackref,breaklinks,colorlinks,allcolors=wacvblue]{hyperref}

\def\confName{WACV}
\def\confYear{2026}

\title{TCDSG: An End-to-End Approach for Action Tracklet Generation}

\author{Raphael Ruschel\\
UC Santa Barbara\\
{\tt\small raphael251@ucsb.edu}
\and
Md Awsafur Rahman\\
UC Santa Barbara\\
{\tt\small awsaf@ucsb.edu}
\and
Hardik Prajapati \\
UC Santa Barbara \\
{\tt\small hprajapati@ucsb.edu}
\and
Suya You \\
ARL West \\
{\tt\small suya.you.civ@army.mil}
\and
B. S. Manjunath \\
UC Santa Barbara \\
{\tt\small manj@ucsb.edu}
}

\newif\ifdraft
\drafttrue  % Uncomment for draft mode
\ifdraft
    \newcommand{\revst}[2]{{\sout{#1} \color{red} #2}}
\else
    \newcommand{\revst}[2]{#2} % Throw away the crossed text and remove the red
\fi

\newcommand{\ignore}[1]{}
\begin{document}
\maketitle
\begin{abstract}
Comprehensive video scene understanding centers on modeling the \emph{temporal evolution of object interactions} across frames. Methods that operate framewise often fragment tracklets and break subject--object linkages, limiting long-term activity analysis. We present \textbf{Temporally Consistent Dynamic Scene Graphs (TCDSG)}, a unified end-to-end framework that jointly performs detection, tracking, and interaction linking across video sequences.
TCDSG is driven by two key ideas. First, a \emph{temporal bipartite matching} strategy maintains stable query assignments across frames, substantially reducing tracklet fragmentation without post-processing. Second, \emph{adaptive decoder queries} augmented with inter-frame feedback inject temporal context directly into decoding, yielding more stable and accurate predictions. As a result, TCDSG retains competitive single-frame accuracy while substantially boosting temporal consistency; for example, \emph{tR@50} improves from 18.6\% to 39.1\% on Action Genome %(+110\% relative).
We evaluate on \textbf{Action Genome}, \textbf{OpenPVSG}, and \textbf{MEVA}, demonstrating consistent gains in spatio-temporal interaction tracking. To support rigorous long-range evaluation, we also re-annotate a subset of MEVA with \emph{persistent cross-frame object IDs}, addressing inconsistencies in the original annotations. The simplicity, robustness, and temporal fidelity of TCDSG make it well suited for continuous video analytics in surveillance, autonomous systems, and human--computer interaction.
\end{abstract}    
\section{Introduction}
\label{sec:intro}
Understanding actions in video underpins applications in large-scale monitoring~\cite{10.1145/3356250.3360041}, human--computer interaction~\cite{robotic,HCI}, and autonomous systems~\cite{bai2023vision}. Scene graphs provide a structured way to encode object relationships, but most work remains \emph{frame-centric}: associations are recomputed independently per frame, so subject--object links rarely persist over time. This gap limits long-range interaction tracking and motivates temporally consistent scene-graph models. 

\begin{figure*}[t]
    \centering
    \includegraphics[width=\linewidth]{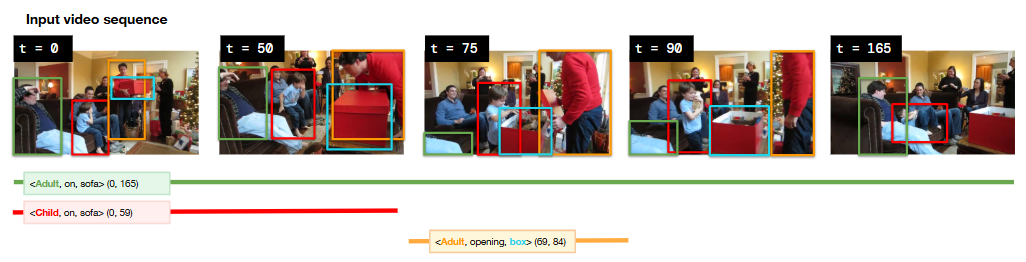}
    \caption{An example of the desired output for a video sequence. At the top, we show a few selected frames and their corresponding timestamp, and at the bottom, distinct action tracklets. Each interaction displays a triplet \(\langle\)subject, object, relationship\(\rangle\) alongside the associated bounding box (color-coded) and timestamps. For clarity, only select interactions are annotated.}
    \label{fig:sample}
\end{figure*}

Figure \ref{fig:sample} presents an example of an input video sequence and the corresponding desired output that includes a triplet $\langle$ \textit{subject, object, relationship} $\rangle$, along with bounding boxes and timestamps to denote the activity.

Recent advances in action detection and scene graph generation have made impressive progress in modeling spatial relationships within frames \cite{hou2021detecting,wang2022learning,hou2022scl}. While some methods address challenges like compositionality \cite{iftekhar2023ddsdecoupleddynamicscenegraph} and dataset biases \cite{TEMPURA}, they approach video understanding as essentially a series of independent frame analyses. Current temporal aggregation strategies typically rely on post-processing to connect frame-level predictions \cite{yang2023pvsg, li2022video}, which introduces several critical limitations:

\begin{enumerate}
    \item Post-processing methods cannot leverage inter-frame information during the detection stage itself
    \item Object identities may swap in crowded scenes
    \item Heuristic linking requires fine-tuning and reduces generalization across datasets
\end{enumerate}

This fundamental disconnect between detection and tracking creates fragmented action tracklets that fail to capture the continuous nature of real-world activities. What's needed is not merely better post-processing, but an approach that inherently maintains temporal consistency throughout the detection process.
To tackle this challenge, we introduce TCDSG: Temporally Consistent Dynamic Scene Graphs, an innovative end-to-end pipeline designed to generate seamless action tracklets across entire video sequences. By integrating our approach into existing architectures with minimal modifications, we achieve substantial gains in temporal consistency for both object tracking and relationship prediction, enabling more robust and continuous video analysis.

We evaluate TCDSG on three benchmark datasets: Action Genome, OpenPVSG, and MEVA, demonstrating state-of-the-art performance in tracklet consistency and competitive framewise prediction. Additionally, we provide a new set of annotations for a subset of the MEVA dataset, where we uniquely identify each object, further validating our approach. Our main contributions include:
\begin{itemize}[leftmargin=1em]
    \item \textbf{Sequence-Level Objective for Temporally Coherent Assignment}: We propose a paradigm shift by redefining the matching objective to operate across time. Instead of optimizing over per-frame entities, we optimize over evolving data trajectories, persistently binding triplets to decoder queries and penalizing identity switching. This sequence-aware formulation is implemented as a lightweight extension of Hungarian matching and enables temporally aligned predictions directly from training.

    \item \textbf{Temporally Conditioned Decoder Queries}:  We design decoder queries that evolve contextually with the video by integrating outputs from previous frames through a cross-attentional feedback loop. This equips the model with memory, temporal context, and continuity, reducing improving prediction stability over time.
 
    \item \textbf{Enhanced MEVA annotations}:% for extended video analytics}: 
    We extend a subset of the MEVA dataset with unique, persistent object IDs, resolving the issue of inconsistent or missing annotations in high-traffic scenes. Although this subset is not the entire dataset, these refined labels enables advanced tracklet generation over longer videos and set a foundation for future work in large-scale dynamic scene understanding.

\end{itemize}

\section{Related works}
\label{sec:related}

\subsection{Scene Graph Generation}

Scene graph generation (SGG) involves creating a graph-like structure where nodes represent objects in a scene and edges represent relationships between these objects, such as actions (e.g., \textit{holding}, \textit{sitting}, \textit{walking}) and spatial relationships (e.g., \textit{close}, \textit{to the right of}, \textit{above}). Originally introduced for image retrieval \cite{johnson2015image}, SGG has since become foundational for applications in monitoring, interactive systems, and human-computer interfaces. Early work focused on \textit{static SGG}, generating scene graphs for single images \cite{lu2021context, tang2019learning, cong2020nodis, wang2019exploring, shi2021simple, wang2021topic, chen2019counterfactual, chiou2021recovering, ulutan2020vsgnet}.

Considerable effort has been devoted to tackling challenges in SGG, such as compositionality and dataset biases. For instance, \cite{iftekhar2023ddsdecoupleddynamicscenegraph} examines compositional SGG to improve recall for unseen triplets formed from known objects and relationships, and also discusses limitations of the two-stage SGG pipeline used in \cite{li2019transferable, liuamplifying, wan2019pose, wang2019deep, gao2020drg, ulutan2020vsgnet}. Additionally, addressing biases from long-tailed distributions is an ongoing focus \cite{VsCGG, TEMPURA}.

Recent advancements have expanded  SGG to dynamic scenes \cite{ji2021detecting, teng2021target, zhang2019graphical, cong2021spatial, Lin2024TD2NetTD}, 
aiming to generate scene graphs across video sequences by harnessing temporal context. Unlike static SGG, dynamic SGG captures time-dependent actions (e.g., \textit{picking up} vs. \textit{putting down}), which require an understanding of temporal dependencies. Temporal consistency becomes essential here, as it supports the accurate interpretation of activities that evolve over time.

An important development in this field is Panoptic Video Scene-Graph Generation (PVSG) \cite{yang2023pvsg}, which introduces action tracklets for temporal activity understanding. This work also proposes the use of panoptic segmentation masks instead of bounding boxes, achieving pixel-level localization for subjects and objects. However, this approach depends on post-processing to aggregate results over time, limiting its utility for applications that require temporal consistency in an end-to-end framework.

\subsection{Tracking}

Object tracking in computer vision focuses on detecting and following objects throughout video sequences to capture their trajectories, a capability essential for applications like autonomous driving, surveillance, action recognition, and augmented reality.  Traditional approaches, including Kalman filters \cite{welch1995introduction} and mean-shift algorithms \cite{zhou2009object}, have evolved with deep learning to form sophisticated multi-object tracking (MOT) systems based on CNNs and transformers.

Many approaches \cite{bewley2016simple, leal2016learning, wojke2017simple, schulter2017deep} address tracking as a combination of appearance and motion cues. However, they rely on post-processing to merge appearance and motion, which often results in suboptimal temporal dependency modeling across frames, limiting its effectiveness for tracking continuous actions. TransTrack \cite{sun2020transtrack}, for example, uses a transformer-based approach for bounding box generation but requires post-processing for IoU-based matching.

While TransTrack leverages transformers for detection, its reliance on post-processing for association contrasts with truly end-to-end approaches like MOTR \cite{zeng2022motr} and TrackFormer \cite{meinhardt2022trackformer} which integrate detection and association directly within their transformer architectures. TrackFormer extend DETR \cite{carion2020end} by treating tracking as a \textit{tracking-by-attention} task, using attention maps for both tracking and detection. MOTR’s tracklet-aware label assignment scheme manages new and stale tracklets dynamically through variable-length track queries.

Despite significant progress, existing methods fall short of unifying activity recognition and object tracking within a fully end-to-end framework. Current approachestreat these tasks in isolation, missing the opportunity to leverage their synergy. TCDSG precisely addresses this critical void by simultaneously capturing complex interactions and object trajectories, which is crucial for advancing applications like trajectory analysis, seamless human-computer interaction, and real-time monitoring in dynamic environments.

\section{Methodology}
\label{sec:method}

In this section, we present our Temporally Consistent Dynamic Scene Graphs (TCDSG) framework for generating coherent action tracklets from video sequences. We first formulate the problem and then describe the architecture, temporal bipartite matching, loss functions, and evaluation metrics. We conclude with a pseudocode outline summarizing the key components of our approach.

\subsection{Problem Formulation}

Given a video sequence $V = \{I_1, I_2, \dots, I_T\}$ of $T$ frames, the goal is to predict a set of action tracklets $\mathcal{AT} = \{at_1, at_2, \dots, at_M\}$, where $M$ is the number of distinct actions occurring in $V$. Each action tracklet $at_i$ is represented as:
\[
at_i = \langle s_i, o_i, r_{s,o}, \mathcal{B}_s, \mathcal{B}_o, t_{start}, t_{end} \rangle
\]
where:
\begin{itemize}
    \item $s_i$ and $o_i$ denote the subject and object class labels,
    \item $r_{s,o}$ is the relationship predicate between subject and object,
    \item $\mathcal{B}_s$ and $\mathcal{B}_o$ are sequences of bounding boxes for subject and object from $t_{start}$ to $t_{end}$,
    \item $t_{start}$ and $t_{end}$ are the start and end frame indices of the tracklet.
\end{itemize}
Our objective is to generate temporally consistent predictions that maintain coherent identities and relations across the entire video duration.

\subsection{Network Architecture}

\begin{figure*}[t]
    \centering
    \includegraphics[width=0.8\linewidth]{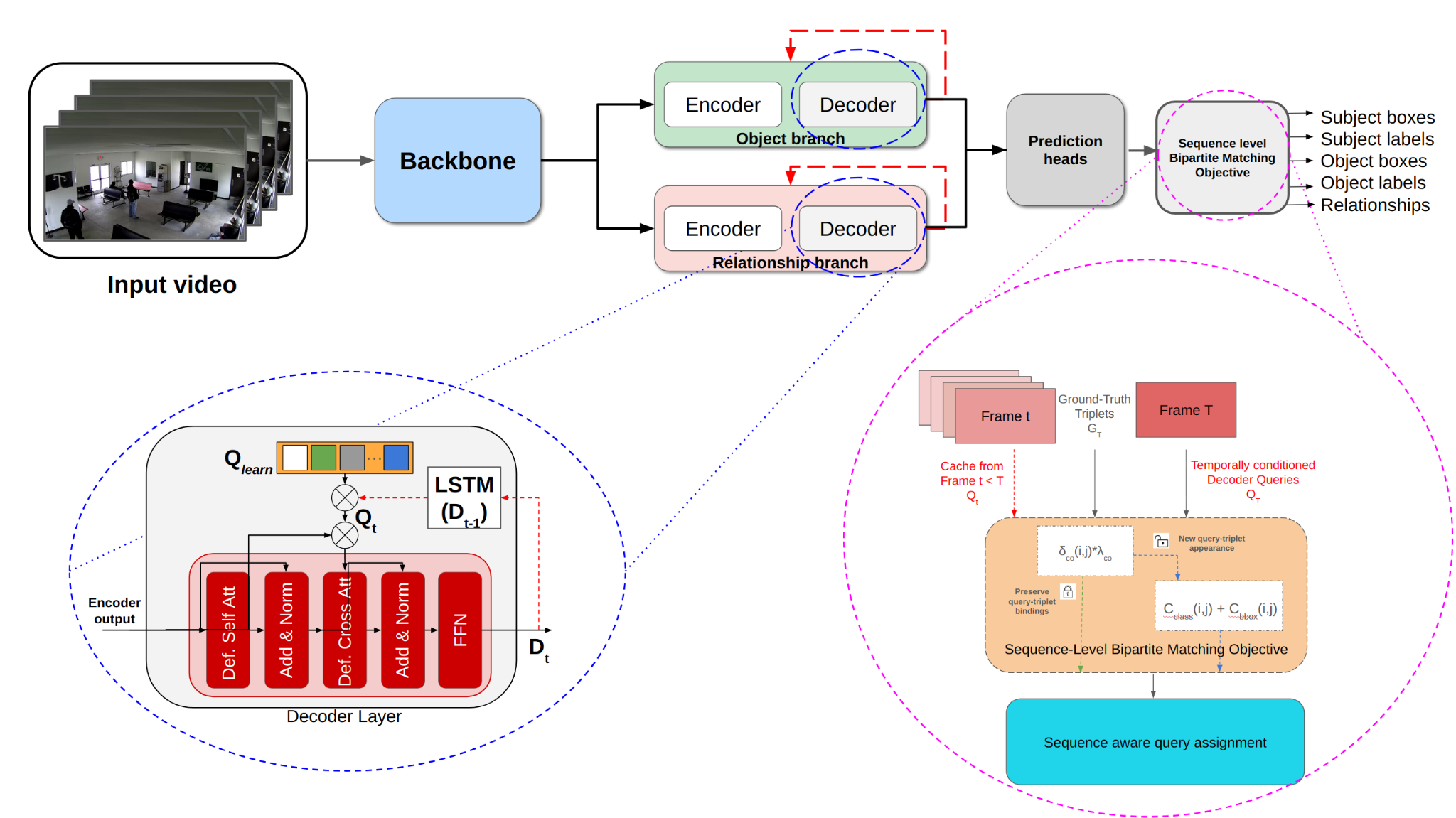}
    \caption{Overview of the TCDSG architecture. A shared CNN backbone extracts features that are fed into two parallel transformer decoder branches for subjects, objects, and relationships. The architecture integrates two key innovations: \textbf{(1) Temporally Conditioned Decoder Queries} (lower left zoomed-in block) that incorporate LSTM-based feedback and cross-attentional context from prior frames, and \textbf{(2) the Sequence-Level Bipartite Matching Objective} (lower right zoomed-in block), which modifies Hungarian matching with a temporal penalty to preserve query-triplet identities across frames. Feedback connections propagate temporal information, enabling end-to-end temporally coherent predictions.}
    \label{fig:architecture}
\end{figure*}

Our architecture builds upon transformer-based detection frameworks such as Deformable DETR~\cite{zhu2020deformable} and DDS~\cite{iftekhar2023ddsdecoupleddynamicscenegraph}, with key modifications to model temporal consistency explicitly. A diagram of our full network is shown in Figure \ref{fig:architecture}.

\textbf{Backbone and Feature Extraction:}
We employ a ResNet-50 backbone~\cite{he2016deep} to extract multi-scale feature maps shared across two parallel branches:
\begin{itemize}
    \item \textbf{Object branch}: predicts object classes and bounding boxes, we consider every unique class in the dataset as possible \textit{subjects} and \textit{objects}.
    \item \textbf{Relationship branch}: predicts predicates between subject-object pairs.
\end{itemize}

\subsection{Temporally Conditioned Queries (TCQ)}
DETR-style transformers rely on static learnable queries, which lack temporal awareness. We introduce \emph{temporally conditioned decoder queries} that evolve by incorporating outputs from prior frames. Specifically, the input queries at frame $t$ are a function of both learnable embeddings and the output embeddings from frame $t-1$, enabling the decoder to adaptively refine its predictions with temporal context.
Previous approaches that incorporates feedback from the previous stages, such as \cite{iftekhar2023ddsdecoupleddynamicscenegraph}, typically just feed the output from the previous frame as input to the current one, however, this presents a shortcoming since the the output from frame $t-1$ is detached from the computation graph, so no gradients will flow through that path during processing of frame $t$. Despite providing significant improvements, we expand on this approach by adding a simple LSTM layer connecting $t-1$ to $t$, ensuring that we maintain an updated state after each frame gets processed. 

Formally, the query embeddings $\mathbf{Q}_t$ at frame $t$ are computed as shown in Equation \ref{eq:query}:
\begin{equation}
\label{eq:query}
\mathbf{Q}_t = f\left( \mathbf{Q}_{\text{learn}}, \; \mathrm{LSTM}(\mathbf{D}_{t-1}) \right)
\end{equation}

\noindent where $\mathbf{Q}_{learn}$ are learned query embeddings, $LSTM$$\mathbf({D}_{t-1})$ is temporal summary of previous decoder outputs, and $f(a,b)$ denotes the cross-attention operation using $a$ as input queries and $b$ as query and value. This design imparts: \begin{itemize}
    \item \textbf{Memory}: retaining information from prior predictions
    \item \textbf{Context}: leveraging temporal dependencies beyond spatial features
    \item \textbf{Continuity}: ensuring that predictions evolve smoothly with the video’s narrative.
\end{itemize}

In Figure \ref{fig:architecture} we show the architecture of our decoders in more details, illustrating this feedback process in detail.

% \begin{figure}[b]
%     \centering
%     \includegraphics[width=0.9\linewidth]{ICCV2025-Author-Kit-Feb/figs/dec_layer_large.png}
%     \caption{Sample decoder layer of TCDSG. We use the same decoder architecture on both the object and relationship branch. Here $\otimes$ represents the cross-attention operation}
%     \label{fig:decoder}
% \end{figure}

\subsection{Sequence-Level Bipartite Matching Objective (SLBM)}\label{sec:TempBiParMatch}
A core contribution of our method is a \emph{sequence-level bipartite matching objective} that optimizes over evolving data trajectories, thus preserving identity across time: once a ground-truth triplet is assigned to a decoder query, we strongly bias subsequent frames to keep that association. This simple change---a lightweight extension to standard Hungarian matching---greatly reduces tracklet fragmentation and yields temporally stable triplet predictions without post-processing. This mechanism is highlighted in the SLBM module of Figure \ref{fig:architecture}

%\paragraph{Standard Hungarian Matching}\revst{Following DETR~\cite{carion2020end}, we use the Hungarian algorithm to find a one-to-one matching between predicted triplets and ground truth in each frame based on a cost matrix combining classification and bounding box regression losses.}{Given that our network architecture is based on the DETR~\cite{carion2020end} architecture, where the classification problem is modeled as a set-prediction problem where $N_q$ predictions are made per frame, with $N_q > N_{obj}$, where $N_{obj}$ is the number of objects in the scene, there is a need to match the predictions to a single ground-truth during training. To do so, the traditional Hungarian Matching algorithm is widely used, however, despite having being used for tracking applications in the past \cite{shanmuga2015eye} the standard algorithm does not incorporate any sort of temporal information, which means that a ground-truth matched to query $q$ on frame $I_t$, can be matched to any query $q \in N_q$ on frames $I_{t+i}$ without any extra penalty, which leads to the same triplet being predicted by different queries over time.}

\textbf{Standard Hungarian Matching:}
Our architecture follows the DETR formulation~\cite{carion2020end}, in which each frame produces $N_q$ set-prediction slots ($N_q > N_{\text{obj}}$) and a one-to-one assignment between predictions and ground-truth objects is computed via Hungarian matching during training. In its vanilla form, standard Hungarian matching minimizes per-frame assignment costs: 
\begin{equation}
    C_t(i,j) = C_{class}(i,j) + C_{bbox}(i,j) 
\label{eq:cost_matrix_vanilla}
\end{equation} 
While effective for static scenes, this formulation fails to preserve temporal continuity. As a result, a ground-truth instance (or subject--object--relation triplet) assigned to query $q$ at frame $I_t$ may be reassigned to any other query $q' \in \{1,\dots,N_q\}$ at frame $I_{t+\Delta}$ with no penalty, requiring additional post-processing ~\cite{shanmuga2015eye}. Over a sequence, these unconstrained framewise permutations cause the \emph{same} triplet to hop across queries, degrading temporal linkage and fragmenting downstream tracklets.

\textbf{Temporal Consistency Enforcement:}
To impose cross-frame identity, we cache each ground-truth triplet $\langle s,o,r_{s,o}\rangle$ the first time it appears and bind it to the decoder query to which it was matched. For all subsequent frames $t$, the Hungarian cost matrix is modified: any assignment that would re-map this previously locked triplet to a different query receives a large penalty $\lambda_{co}$. This simple cache-and-penalize extension to standard (per-frame) Hungarian matching~\cite{carion2020end,shanmuga2015eye} sharply reduces tracklet fragmentation by preserving stable query$\rightarrow$triplet identities across time. Because the constraint is injected during training, inference inherits temporally aligned query indices, allowing tracklets to be formed by straightforwardly grouping per-frame predictions by query.

%\paragraph{Temporal Consistency Enforcement} We extend this to the temporal domain by maintaining a record of unique triplets $\langle s, o, r_{s,o} \rangle$ and their assigned query indices. When processing frame $t$, the cost matrix is augmented by imposing a large penalty on query assignments that conflict with previously locked triplets. This ensures that once a triplet is assigned a query at an earlier frame, future frames must respect this assignment, thereby producing temporally coherent tracklets.

Formally, for prediction $i$ and ground truth $j$ at frame $t$, the cost matrix $C_t$ is as defined in \ref{eq:cost_matrix}:
\begin{equation}
    C_t(i,j) = C_{class}(i,j) + C_{bbox}(i,j) + \delta_{co}(i,j) \cdot \lambda_{co}
\label{eq:cost_matrix}
\end{equation}

Where $C_{class}(i,j)$ is the classification cost between query $i$ and ground-truth label $j$, similarly, $C_{box}(i,j)$ is the cost to match bounding box $i$ to the $j$th ground-truth box. Lastly $\delta_{co}(i,j)$ is 1 if assignment conflicts with prior frames, and $\lambda_{co}$ is a positive, large scalar. A pseudocode of the forward pass of our network during training is shown in Algorithm \ref{alg:temporal_matching}.

\begin{algorithm}
\caption{TCDSG forward pass}
\label{alg:temporal_matching}
\begin{algorithmic}[]
\State \textbf{Input:} Video frames $\{I_1, \dots, I_T\}$, learnable queries $\mathbf{Q}_{learn}$
\State Initialize hashmap $\mathcal{H} \gets \emptyset$ to store triplet-query assignments
\For{$t=1$ to $T$}
    \State Extract features $\mathbf{F}_t$ from $I_t$ via backbone
    \If{$t=1$}
        \State Set query embeddings $\mathbf{Q}_t \gets \mathbf{Q}_{learn}$
    \Else
        \State Obtain previous decoder outputs $\mathbf{D}_{t-1}$
        \State Update queries $\mathbf{Q}_t$ as in Eq. \ref{eq:query}
    \EndIf
    \State Decode predictions $\mathcal{P}_t$ from $\mathbf{F}_t$ and $\mathbf{Q}_t$
    \State Compute cost matrix $C_t$
    \ForAll{triplets in $\mathcal{H}$}
        \State Add large penalty to corresponding entries in $C_t$
    \EndFor
    \State Perform Hungarian matching on $C_t$
    \ForAll{$(p_i, g_j) \in M_t$}
        \If{$g_j$ not in $\mathcal{H}$}
            \State Lock assignment: $\mathcal{H}[g_j] \gets p_i$
        \EndIf
    \EndFor
    \State Compute losses and update model parameters
\EndFor
\end{algorithmic}
\end{algorithm}

\subsection{Loss Functions}

Our total loss $\mathcal{L}$ combines several terms balancing spatial accuracy, semantic consistency, and temporal stability, and is defined in Eq.~\ref{eq:total_loss}:

\begin{equation}
    \mathcal{L} = \lambda_{spat} \mathcal{L}_{spat} + \lambda_{const} \mathcal{L}_{const} + \lambda_{ref} \mathcal{L}_{ref} + \lambda_{label} \mathcal{L}_{label}
    \label{eq:total_loss}
\end{equation}

\begin{itemize}
    \item \textbf{Spatial loss} $\mathcal{L}_{spat}$: combines generalized IoU and $L_1$ regression of the bounding box normalized coordinates.
    \item \textbf{Label Loss} $\mathcal{L}_{label}$: Cross-entropy for class labels, and focal loss for predicates as in prior work~\cite{carion2020end}.
    \item \textbf{Consistency loss} $\mathcal{L}_{const}$: encourages feature embeddings of the same class across frames to be similar, improving temporal feature stability.
    \item \textbf{Reference point loss} $\mathcal{L}_{ref}$: The Reference point loss  guides the initial reference points for deformable attention, typically towards object centers, to stabilize early-frame predictions when dynamic query feedback from previous frames is not yet available (t=1)
\end{itemize}

% Where $\lambda_{i}$ is a weight factor for the corresponding loss term that is assigned empirically, where we found that setting $\lambda_{spat}=2.5$, and all others at 1 yield the best result.

\begin{table*}[t]
\centering
\begin{tabular}{@{}lcccccc@{}}
\toprule
& \multicolumn{6}{c}{SGDet - With constraint} \\
\cmidrule(lr){2-7}

Method & mR@20 & mR@50 & R@20 & R@50 & \textbf{tR@20} & \textbf{tR@50} \\ 
\midrule
iSGG \cite{khandelwal2022iterative} & 19.7 & 22.9 & 29.2 & 35.3 & - & - \\
STTran-TPI \cite{STTran-TPI} & 20.2 & 21.8 & 29.1 & 34.6 & - & - \\
APT \cite{li2022dynamic} & - & - & 29.1 & 38.3 & - & - \\
TEMPURA \cite{TEMPURA} & 22.6 & 23.7 & 33.4 & 34.9 & - & - \\
VsCGG \cite{VsCGG} & - & 24.2 & 35.8 & 38.2 & - & - \\
TD2-Net (p) \cite{Lin2024TD2NetTD} & - & 23 & 28.7 & 37.1 & - & - \\
DSG-DETR \cite{feng2023exploiting} & - & - & 34.8 & 36.1 & - & - \\
OED \cite{wang2024oed} & - & - & 40.9 & 48.9 & - & - \\
DDS \cite{iftekhar2023ddsdecoupleddynamicscenegraph} & 29.1 & 32.2 & 42.0 & 47.3 & 13.5 & 18.6 \\
TPT \cite{zhang2023end} & - & - & 37.3 & 49.2 & - & - \\
\textbf{TCDSG* (Ours)} & \textbf{36.1} & \textbf{46.8} & \textbf{47.9} & \textbf{58.4} & 14.4 & 27.1 \\
\textbf{TCDSG (Ours)} & 27.8 & \textbf{38.9} & 41.6 & \textbf{52.5} & \textbf{17.5} & \textbf{39.1} \\

\bottomrule
\end{tabular}
% \vspace{0.2cm}
\caption{Comparisons on the Action Genome dataset under the SGDet (with constraint) protocol. While our method (\textbf{TCDSG}) yields competitive \emph{R@k}, it exhibits substantially higher (\emph{tR@k}), indicating its robust ability to capture relationships across longer sequences. For instance, \emph{tR@50} improves from \emph{18.6} to \emph{39.1} in certain settings, highlighting TCDSG’s effectiveness in generating coherent action tracklets over time. \textbf{TCDSG*} refers to our method when we remove the Temporal Hungarian Matching during training, showing the tradeoff between frame-level and temporal-based recall.}
\label{tab:AG_results}
\end{table*}

\subsection{Evaluation Metrics}

We adopt temporal Recall@K ($tR@K$) to measure tracklet consistency. Unlike frame-level Recall@K, $tR@K$ requires predictions to have both spatial overlap (IoU $\geq 0.5$) and sufficient temporal Intersection over Union with ground truth. This metric better reflects performance on continuous action understanding.

\section{Datasets}

To train and evaluate our method, we utilized three datasets that provide bounding box annotations for subjects and objects, along with their inter-relationships. Each dataset was adapted to ensure consistent object identifiers, crucial for robust tracking.

\subsection{Action Genome (AG)}

Action Genome (AG) \cite{ji2020action} extends the Charades dataset \cite{sigurdsson2016hollywood} with frame-level scene graph annotations. AG provides bounding box and relationship annotations for 36 object and 25 relationship classes. However, subjects are exclusively labeled as "person," and unique identifiers for tracking are not provided.

To enable tracking, we implemented a pseudo-labeling strategy to assign consistent object identifiers across frames based on bounding box overlap. While this approach may introduce noise if distinct objects of the same class frequently enter and exit the scene, our manual verification indicated that such instances typically involved the same object re-entering, validating our methodology's applicability.

\subsection{OpenPVSG}

The OpenPVSG dataset \cite{yang2023pvsg} was developed for Panoptic Video Scene Graph Generation (PVSG), which extends traditional scene graph generation with pixel-level segmentation masks. Comprising 400 videos (averaging 76.5 seconds at 5 FPS) sourced from VidOR \cite{shang2019annotating}, EpicKitchen \cite{Damen2018EPICKITCHENS}, and Ego4D \cite{grauman2022ego4d}, OpenPVSG offers diverse indoor and outdoor scenes captured by both moving and static cameras. Unlike AG, OpenPVSG includes various subject types beyond "person" and provides unique object identifiers per video sequence, making it inherently suitable for tracking tasks.

OpenPVSG features 157 object classes (e.g., animals, furniture, food) and 57 relationship classes. For compatibility with our methodology, we converted the provided segmentation masks into bounding boxes by extracting the minimum and maximum coordinates.

\subsection{MEVA}

The MEVA dataset \cite{Corona_2021_WACV} is a large-scale collection for activity detection in multi-camera environments.  %was developed under the IARPA DIVA program. 
It comprises over 9300 hours of untrimmed video footage. Our study utilized 144 hours of annotated data covering 37 activity types, with corresponding bounding boxes for each instance. This data was collected from approximately 100 actors performing scripted scenarios over three weeks at an access-controlled venue, captured by 38 cameras configured for typical surveillance setups.

Despite its scale, MEVA presents several annotation challenges:
\begin{itemize}
    \item Some relationships lack a standard \texttt{<subject, object, relationship>} triplet format (e.g., "vehicle reverses" without an explicit object).
    \item Certain relationship classes embed objects within their labels (e.g., "person opens facility door"), but the corresponding objects may lack bounding box annotations.
    \item Intended unique track IDs often result in multiple IDs for the same individual within a single video sequence, compromising tracking reliability.
\end{itemize}

We therefore re-annotated a subset of MEVA with unique, cross-frame object IDs, enabling consistent multi-frame evaluation under our R@K / tR@K metrics. Our additional annotations for MEVA include 141 training videos ($\approx400k$ frames across 16 cameras) and 47 test videos ($\approx23k$ frames across 12 cameras, 3 of which were unseen during training). We focused on re-annotating the busiest regions of the dataset, such as the bus station, subway station, and the regions surrounding them.

\subsection{Implementation Details}
We use a ResNet-50~\cite{he2016deep} backbone and train with AdamW~\cite{ilya2019decoupled} using BLoad~\cite{ruschel2023bload} for efficient, distributed learning on variable-length videos.

Each branch uses a 6-layer encoder--decoder stack with $N_q{=}100$ 256-d queries. We initialize from DETR~\cite{carion2020end} weights pre-trained on COCO~\cite{lin2014microsoft} and train for 20 epochs on $8\times$NVIDIA A100 GPUs, selecting the best checkpoint by validation metrics. The full model has 59M parameters and runs at $\sim$20\,fps using $<2$\,GB GPU memory at inference (no optimizations).

\textbf{Inference \& Tracklet Construction:}
During inference, TCDSG directly outputs coherent tracklets without post-hoc linking, no additional IoU linking, or ReID post-processing is required. Thanks to our temporal matching (Sec.~\ref{sec:TempBiParMatch}), predictions maintain consistent query-triplet associations and are grouped by query index. For each frame we compute a joint confidence score per prediction, rank, and retain the top-$k$ entries (consistent with Recall@$k$ evaluation). Tracklets are formed by concatenating consecutive frames whose predictions share the same query index \emph{and} triplet label.

This native mapping yields tracklets that are temporally trimmed and identity-stable: a given query index corresponds to a single evolving $\langle s,o,r_{s,o}\rangle$ interaction for the duration of its visibility. If a previously active query drops from the top-$k$ set or changes its predicted triplet, we close the current tracklet and (if re-emitted) initialize a new one, allowing graceful handling of occlusion, interaction changes, or confidence decay. Integrating detection, relationship prediction, and temporal linking within one transformer-based pipeline is a key differentiator of TCDSG compared with methods that must post-hoc stitch per-frame scene graphs into action tracklets.

\section{Results}
\subsection{Performance evaluation}
\begin{figure*}[t]
    \centering
    \includegraphics[width=0.95\linewidth]{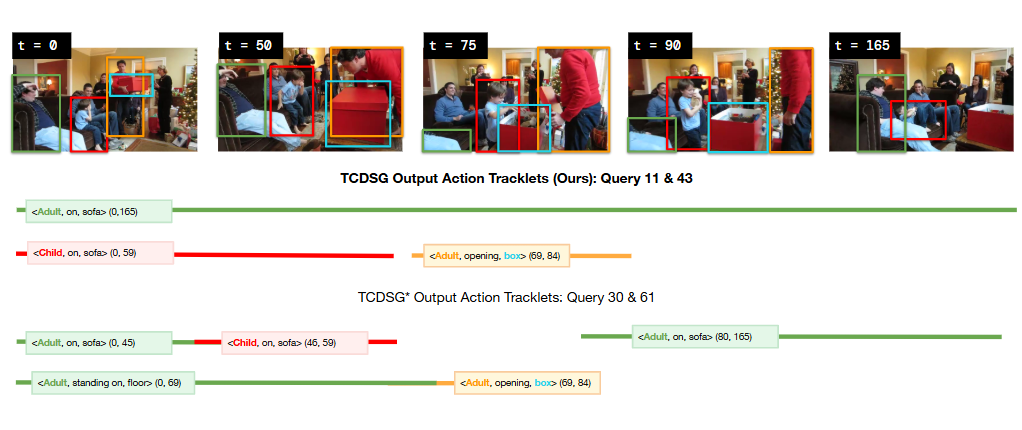}
    \caption{Comparative visualization of action tracklet continuity between TCDSG (our method) and TCDSG*, which represents our module without the proposed temporal matching. While TCDSG* produces fragmented tracklets with the same query producing different triplets, TCDSG maintains consistent query assignment throughout the video sequence. This persistent identity mapping enables continuous tracking of interactions over extended durations. For clarity, only boxes from TCDSG are plotted}
    \label{fig:qual}
\end{figure*}

We evaluate TCDSG on Action Genome against state-of-the-art scene-graph / HOI baselines. Table~\ref{tab:AG_results} aggregates reported results; metrics absent from original papers (e.g., tR@K) are shown as ``--''. Figure~\ref{fig:qual} illustrates a representative qualitative comparison, highlighting the improved temporal coherence of our predictions. Additional qualitative results are included in the supplementary material.

Despite incomplete metric coverage across baselines, TCDSG delivers competitive single-frame Recall@K and \emph{substantially} higher temporal scores. As shown in Table~\ref{tab:AG_results}, it achieves the strongest tR@K, demonstrating stable subject--object relationship tracking over extended sequences. These gains follow from our temporally aware design—temporal bipartite matching and inter-frame feedback—which curbs query permutation and tracklet fragmentation common to framewise pipelines. Such long-horizon stability is critical for continuous video analytics in surveillance, monitoring, and activity understanding.

Although we evaluate OpenPVSG in \emph{bounding-box mode} (converting its panoptic masks), it remains markedly more challenging than AG: clips are longer, object and predicate vocabularies are larger, and subjects span many categories (AG: subject always \texttt{person}), adding an additional label dimension to the matching problem. Accordingly, absolute recall values in Table~\ref{tab:pvsg} are lower than on AG. Even so, TCDSG yields improved temporal recall, underscoring its robustness under greater semantic and temporal variability.

\begin{table}[h]
\centering
\setlength{\tabcolsep}{5pt}
\begin{tabular}{lcccc}
\toprule
Method             & R@20 & R@50 & \textbf{tR@20} & \textbf{tR@50} \\
\midrule
IPS + T \cite{cheng2022masked, wang2021different}          &  -    &  -    &  3.88     &  5.66     \\
VPS \cite{cheng2022masked, li2022video}               & - & - & 0.42 & 0.73 \\
% \textbf{TCDSG* (baseline)}    &  \textbf{32.7 }   &  \textbf{36.6 }   &   4.0    &  4.5     \\
MACL \cite{nguyen2025motion} & - & - & 4.5 & 6.0 \\
\textbf{TCDSG*} &   \textbf{27.5}   &  \textbf{32.3}    &   \textbf{10.2}    &  \textbf{14.2}    \\
\bottomrule
\end{tabular}
\caption{Results on OpenPVSG dataset. For IPS+T and VPS results, we extracted the numbers from \cite{yang2023pvsg}. On our results, $*$ denotes using bounding boxes as object location grounding.}
\label{tab:pvsg}
\end{table}

% \st{Similar to AG, we note a significant improvement in the temporal recall@K over previous methods, highlighting the advantages of our method.}

%\rev{In the MEVA dataset, participants in the DIVA program, such as \cite{dave2022gabriellav2}, evaluated their methods using different metrics, primarily Pmiss@Xtfa. Here, Pmiss represents the ratio of activities where the system did not detect the activity for at least one second, while TFA (time-based false alarm rate) measures the portion of time that the system incorrectly detected an activity. Due to differences in evaluation metrics and the fact that we use only a subset of MEVA data with additional annotations from our tool, direct comparisons with DIVA participants are not feasible.}

%Given that MEVA includes videos up to five minutes long, we leverage this unique property to test our method’s robustness across various video lengths by introducing different subsampling factors for the test data. This approach allows us to assess temporal consistency in conditions ranging from densely sampled sequences to more sparsely sampled ones. Results are shown in Table \ref{tab:meva}.

\textbf{MEVA Evaluation:}
MEVA poses additional challenges: most prior work (e.g., \cite{dave2022gabriellav2}) reports \emph{Pmiss@Xtfa} rather than Recall@K. In addition, we operate on the newly curated subset and with different metrics, direct comparison to prior MEVA reports is not meaningful.

MEVA’s long clips (up to $\sim$5\,min) provide a rigorous stress test for temporal stability. We observed large gaps between framewise and temporal recall, motivating an analysis of \emph{temporal sampling density}. In Table~\ref{tab:meva} we evaluate multiple subsampling factors (keeping every $f^{\text{th}}$ frame). R@K remains stable, but tR@K improves markedly as $f$ increases, suggesting that long unbroken sequences expose transient query switches that fragment tracklets; sparser sampling reduces these opportunities. Going forward, we plan to incorporate short-term multi-hypothesis or temporally smoothed assignment strategies to further suppress such transient switches in long-form video.

% \bsm {Despite the new annotations, direct comparison with \bsm{DIVA participants?? or papers} remains infeasible because we evaluate on Recall@K and temporal Recall@K (tR@K), and only for a select subset. \textbf{I dont like this statement and not sure if this is accurate either, but put this in to have a discussion on why direct comparison is not possible.} Nonetheless, MEVA’s longer clips (up to five minutes) provide a rigorous test bed for our method’s ability to maintain coherent tracklets across thousands of frames.  In Table~\ref{tab:meva}, we vary the \emph{subsampling factor} to assess how frame spacing influences temporal consistency. Higher subsampling tends to improve tR@K, suggesting fewer opportunities for transient \"switches\" between tracklets. }

% Still, two limitations emerge from Table~\ref{tab:meva}: (1) our current reliance on single query indices can fragment a continuous activity if a query briefly shifts from triplet $T_1$ to $T_2$ and back again, and (2) densely populated scenes can trigger "track switching" when multiple visually similar subjects cause the decoder to drift. Future work will incorporate identity-verification modules or context-aware reassignments, further enhancing tracklet stability in large-scale, high-density environments typical of MEVA.

\begin{table}[h]
\centering
\setlength{\tabcolsep}{5pt}
\begin{tabular}{lcccc}
\toprule
Subsample Factor             & R@20 & R@50 & \textbf{tR@20} & \textbf{tR@50} \\
\midrule
1 & 62.1 & 67.9 & 6.7 & 7.8 \\
5 & 62.1 & 67.9 & 9.9 & 12.4 \\
10 & 62 & 67.7 & 10.9 & 17 \\
20 & 61.9 & 67.6 & 17.6 & 31.2 \\
\bottomrule
\end{tabular}
\caption{Results on MEVA dataset with different subsampling factors on our proposed test set}
\label{tab:meva}
\end{table}

%As seen in Table \ref{tab:meva}, temporal Recall@K (tR@K) generally increases with higher subsampling factors. However, this trend highlights a key limitation: stacking predictions solely based on query index can lead to prediction fragmentation due to noise. For example, if a query predicts a triplet $T_1$ consistently over several frames but briefly switches to $T_2$ before returning to $T_1$, this results in three separate tracklets, disrupting continuity. This issue is particularly evident at lower subsampling factors (1 and 5) on longer videos, where actions may span thousands of frames and encounter higher noise levels.

%Another limitation is track switching in densely populated scenes, where multiple subjects perform similar activities. In such cases, a query index may briefly switch the tracked subject, leading to failed spatial IoU checks and inaccurate tracking. Addressing these limitations could involve adding identity verification mechanisms or context-aware adjustments to enhance tracklet stability and reduce fragmentation in challenging, high-density environments like those in MEVA.

\section{Ablation Studies}
We ablate TCDSG components on Action Genome (default dataset unless noted). Table~\ref{tab:ab_AG} reports incremental additions to the baseline and their impact on frame-level Recall@K (R@K) and temporal Recall@K (tR@K).

%In this section, we conduct ablation studies to evaluate the impact of each component of our proposed approach. Unless otherwise noted, all experiments were conducted on the Action Genome dataset, which provides a balanced combination of class diversity and data availability.

%Table \ref{tab:ab_AG} shows the performance of our baseline model and the effect of adding each component on both single-frame predictions (R@k) and activity tracking (tR@k).

\begin{table}[h]
\centering
\setlength{\tabcolsep}{1.5pt}
\begin{tabular}{lcccc}
\toprule
\textbf{Method} & R@20 & R@50 & \textbf{tR@20} & \textbf{tR@50} \\ \hline
(1) Baseline & 36.7 & 44.4 & 10.2 & 16.0 \\ 
(2) SLBM & 34.3 & 42.0 & 13.1 & 25.4 \\ 
(3) TCQ & 39.8 & 46.9 & 11.0 & 16.2 \\
(4) Cos similarity + (2) + (3) & 38 & 47.8 & 15.5 & 30.2 \\
(5) $\mathcal{L}_{ref}$ + (4) & 39.7 & 51.0 & 16.5 & 35.6 \\
(6) LSTM + (5) &  \textbf{41.6} & \textbf{52.5} & \textbf{17.5} & \textbf{39.1} \\
(7) (6) -- (2) & \textbf{47.9} & \textbf{58.4} & 14.4 & 27.1 \\
\bottomrule
\end{tabular}
\caption{Performance comparison of our baseline model and the effect of each proposed component on R@k and tR@k at different $k$ values on the Action Genome dataset.}
\label{tab:ab_AG}
\end{table}

\textbf{Baseline (No Temporal Locking):}
When run without temporal consistency enforcement, our network delivers \emph{competitive} state-of-the-art frame-level Recall@K (R@K) on Action Genome, underscoring the strength of the underlying transformer architecture and adaptive query design for single-frame scene graph generation.

\textbf{Adding Temporal Consistency:}
Introducing temporal bipartite matching---which preserves query identity across frames---can slightly reduce per-frame R@K. This is an intentional trade-off: once a query is “claimed” by a triplet, we discourage reassignments that might improve instantaneous detection at the cost of long-range coherence.

The benefit of enforcing identity persistence is clear in temporal metrics. As shown in Table~\ref{tab:AG_results}, TCDSG increases tR@50 from 18.6\% (DDS) to 39.1\% (\,+110\% rel.\,), demonstrating substantially more stable subject--object relationship tracking over time. These results validate our end-to-end, temporally grounded formulation as a practical route to robust action tracklet generation.

Integrating cosine similarity significantly boosts temporal recall by enforcing consistency in feature representations across consecutive frames. This aligns with the understanding that features for recurring objects or interactions should remain stable throughout a video, given the inherently smooth temporal transitions. 

The $\mathcal{L}_{ref}$ is crucial for guiding the initial reference points of the deformable attention mechanism. This is especially beneficial in the early frames of a video sequence, where the model lacks sufficient temporal context, and the dynamic queries haven't yet accumulated enough information to effectively guide the model.

Lastly, incorporating an LSTM layer into the feedback loop serves to compute an internal state representation of the previous frame. The LSTM's inherent update and forget gates allow the network to selectively retain only the most informative data from previous queries, thereby refining temporal context and improving overall prediction stability.

\section{Conclusion \& Future Works}

In this paper, we introduced TCDSG, an end-to-end pipeline designed to generate temporally consistent scene graphs, enabling the creation of robust action tracklets that are applicable to various downstream tasks in video-based analytics and surveillance. Our approach introduces two key paradigm shifts for video scene understanding: a sequence-level objective for trajectory-aware query assignment, and temporally conditioned decoder queries that evolve dynamically with the scene. Together, these contributions enable robust action tracklet generation without post-processing, achieving strong temporal consistency across diverse datasets. Through evaluations on three benchmark datasets, TCDSG achieved state-of-the-art results in tracklet predictions while maintaining competitive performance on single-frame predictions.

Additionally, leveraging the rich MEVA dataset, we plan to explore Re-identification (ReID) capabilities across distinct cameras by having an additional ReID block on the network that access a centralized memory bank that contains shared information from all cameras. By focusing on cross-camera ReID, we can extend TCDSG’s capabilities to track individuals across multiple camera views, a feature highly valuable in multi-camera surveillance setups. This enhancement has the potential to enable robust tracking across complex environments, supporting continuous monitoring in large-scale applications.

\newpage

{
    \small
    \bibliographystyle{ieeenat_fullname}

    % \bibliography{main, refs/hoi, refs/generic, refs/vrl-papers, refs/tracking}
}

\end{document}